\documentclass[a4paper,conference]{IEEEtran}
\usepackage{cite}
\ifCLASSINFOpdf
   \usepackage[pdftex]{graphicx}
   \graphicspath{{images/results/84/}{images/results/240/}}
   \DeclareGraphicsExtensions{.pdf,.jpeg,.png}
\else
\fi
\usepackage{amsmath}

\usepackage{stfloats}
\usepackage{url}
\usepackage{amssymb}
\usepackage{pifont}% http://ctan.org/pkg/pifont

\usepackage{enumitem}
\DeclareMathOperator{\argminG}{arg\,min} % Jan Hlavacek
\usepackage{bbold}
\usepackage[T1]{fontenc}
\usepackage{booktabs}
\usepackage{caption}
\usepackage{subcaption}
\usepackage[pagebackref,breaklinks,colorlinks]{hyperref}
\usepackage{xcolor}

\begin{document}
\title{MSF3DDETR: Multi-Sensor Fusion 3D Detection Transformer for Autonomous Driving}

\author{\IEEEauthorblockN{Gopi Krishna Erabati\textsuperscript{*} and Helder Araujo}
\IEEEauthorblockA{Institute of System and Robotics\\
University of Coimbra, Portugal\\
Email: \{gopi.erabati, helder\}@isr.uc.pt}}
% \and
% \IEEEauthorblockN{Homer Simpson}
% \IEEEauthorblockA{Twentieth Century Fox\\
% Email: homer@thesimpsons.com}
% \and
% \IEEEauthorblockN{James Kirk\\ and Montgomery Scott}
% \IEEEauthorblockA{Starfleet Academy\\
% San Francisco, California 96678--2391\\
% Telephone: (800) 555--1212\\


%Telephone: (800) 555--1212, Fax: (888) 555--1212}


\maketitle

\begingroup\renewcommand\thefootnote{*}
\footnotetext{Corresponding author}
\endgroup

\begin{abstract}
    3D object detection is a significant task for autonomous driving. Recently with the progress of vision transformers, the 2D object detection problem is being treated with the set-to-set loss. Inspired by these approaches on 2D object detection and an approach for multi-view 3D object detection DETR3D, we propose MSF3DDETR: Multi-Sensor Fusion 3D Detection Transformer architecture to fuse image and LiDAR features to improve the detection accuracy. Our end-to-end single-stage, anchor-free and NMS-free network takes in multi-view images and LiDAR point clouds and predicts 3D bounding boxes. Firstly, we link the object queries learnt from data to the image and LiDAR features using a novel MSF3DDETR cross-attention block. Secondly, the object queries interacts with each other in multi-head self-attention block. Finally, MSF3DDETR block is repeated for $L$ number of times to refine the object queries. The MSF3DDETR network is trained end-to-end on the nuScenes dataset using Hungarian algorithm based bipartite matching and set-to-set loss inspired by DETR. We present both quantitative and qualitative results which are competitive to the state-of-the-art approaches.
\end{abstract}


% \fi
\IEEEpeerreviewmaketitle

\section{Introduction}
\label{sec:intro}
3D object detection is one of the fundamental and the most crucial element of the visual perception system of autonomous vehicles. Object detection is a twofold problem of classifying and localizing the objects in the scene. A 3D LiDAR point cloud provides the depth information of the scene but the point cloud data is very sparse at long range. On the other hand, a RGB sensor provides dense pixel data of the scene but it does not perceive the depth and it is very sensitive to illumination changes in the scene. Therefore, high-resolution RGB and 3D LiDAR data can be fused to leverage their complementary nature.

The well researched 2D CNN based object detection techniques \cite{ssd, yolo} are not well suited for 3D LiDAR point clouds due to their sparse and unstructured nature. In order to tackle this challenge, researchers employed various approaches such as voxelization or discretization of point clouds. Several approaches \cite{voxnet, second} discretize the point cloud into 3D voxels to apply regular 3D CNNs, but they are computationally expensive. Other approaches \cite{mv3d, avod} uses spherical or cylindrical projection to obtain Birds-Eye View (BEV) representation of point clouds and apply 2D CNNs. In our method we test two approaches: an efficient BEV representation of 3D LiDAR point clouds obtained using PointPillars\cite{pointpillars}, and SparseConv \cite{sparseconv} to obtain 3D sparse convolutions on the voxels.

% Many approaches \cite{mv3d, avod} employ two-stage pipeline for 3D object detection, including region proposal stage which is a bottleneck for optimization during training. To tackle this challenge single-stage anchor-based approaches \cite{pixor, complexyolo} were proposed to ease the training. To remove the dataset specific anchor boxes from the detection pipeline few approaches employed per-pixel prediction. However, these approaches predicts redundant boxes which are removed using post-processing technique, such as non-maximum suppression (NMS). Our approach does not rely on hand crafted post-processing methods like NMS.

We propose Multi-Sensor Fusion 3D DEtection TRansformer (MSF3DDETR) a single-stage, anchor-free and post-processing (NMS) free network for 3D object detection leveraging the fusion of multi-view RGB and LiDAR data. The multi-view RGB features are extracted by a shared ResNet backbone \cite{resnet} and LiDAR features are extracted by SECOND backbone \cite{second} from the VoxelNet \cite{voxelnet} or PointPillars \cite{pointpillars}. We introduce a novel MSF3DDETR cross-attention block which fuses the RGB and LiDAR features by the attention mechanism using the sparse set of object queries which are learnt end-to-end. To our knowledge, this is the first approach to fuse the RGB and LiDAR features by attention mechanism leveraging the learnt object queries. Similar to \cite{detr3d}, the fused features interact with each other through a multi-head self-attention layer \cite{vaswani2017attention}. The 3D bounding box parameters are regressed from every layer of MSF3DDETR block and we use bipartite matching with a set-to-set loss inspired by DETR \cite{detr} to optimize the model during training.

We train our network on the publicly available autonomous driving nuScenes dataset \cite{caesar2020nuscenes} and obtained competitive results (without NMS) to the state-of-the-art methods (with NMS).

\noindent Our main contributions are as follows:
\begin{itemize}[noitemsep,topsep=0pt]
    \item We propose an end-to-end, single-stage, anchor-free and NMS-free MSF3DDETR network to detect 3D objects. Similar to DETR \cite{detr}, our method does not require any post-processing such as NMS, however it obtains competitive results with existing NMS-based approaches.  We also employ knowledge distillation with teacher and student model to improve the accuracy.
    \item Our approach fuses the multi-view RGB and LiDAR features leveraging the learnt object queries using the cross-attention mechanism. To our knowledge, this is the first attempt to fuse multi-view RGB and LiDAR features using a novel cross-attention block at the object query level and posing multi-sensor fusion as a 3D set-to-set prediction.
    \item We release our code and models to facilitate further research.
\end{itemize}

\section{Related Work}
\label{sec:relatedwork}

\textbf{CNN-based object detection} The CNN-based methods span from two-stage approaches to anchor-free approaches. RCNN  \cite{rcnn} and its variants Fast RCNN \cite{fastrcnn} and Faster RCNN \cite{fasterrcnn} are two-stage 2D object detection methods which are typically slow in practice. Point RCNN \cite{shi2019pointrcnn} and Faster Point RCNN \cite{fastpointrcnn} are two-stage 3D object detection approaches which aggregate RoI specific features for object proposals using PointNet set abstraction layer \cite{qi2017pointnet++}. Region proposal network \cite{fasterrcnn} is removed in the single-stage approaches \cite{ssd, yolo}, thus making the detection faster than two-stage approaches. VoxelNet \cite{voxelnet} is a single-stage 3D object detection approach which uses 3D convolutions, which are bottleneck in computation efficiency. Some approaches such as PIXOR \cite{pixor} and PointPillars \cite{pointpillars} project the point cloud to BEV representation and use 2D CNNs to predict 3D bounding boxes. 
% Complex-YOLO \cite{complexyolo} is a single-stage 3D object detection method inspired from YOLO \cite{yolo}. 
However, these methods contain the anchor boxes which are statistically obtained from the dataset. Few approaches like FCOS \cite{tian2019fcos}, CenterNet \cite{duan2019centernet} and CornerNet \cite{law2018cornernet} shift from per-anchor detection to per-pixel or per-keypoint 2D object detection. FCOS3D \cite{wang2021fcos3d} is a monocular 3D object detection approach which employs per-pixel prediction. PillarOD \cite{pillarod} and CenterPoint \cite{centerpoint} are anchor-free 3D object detection approaches which employ per-pillar (BEV pixel) prediction. However, these methods still predict redundant bounding boxes which are removed by post-processing methods, such as NMS.

\textbf{Transformer-based object detection} To remove the post-processing (like NMS), DETR \cite{detr} defines the object detection problem as a set-to-set problem. It uses bipartite matching using Hungarian algorithm to match the predictions with the ground-truth boxes, thus removing the post-processing to remove redundant boxes. To accelerate the training Deformable DETR \cite{zhu2020deformable} employs deformable self-attention. DETR3D \cite{detr3d} and Object DGCNN \cite{objectdgcnn} are 3D object detection approaches based on set-to-set prediction problem. Our method also defines 3D object detection as a set-to-set prediction problem, but with a novel cross-attention block to fuse RGB and LiDAR features.

\textbf{Fusion-based 3D object detection} approaches fuse data of 3D LiDAR and RGB camera. MV3D \cite{mv3d} and AVOD \cite{avod} fuses the projected LiDAR and RGB data using a two-stage CNN approach to predict the 3D bounding boxes. The objects are fused at the proposal stage using RoIPooling. Frustum PointNet \cite{qi2018frustum} uses the 2D object detection on RGB images to narrow the search space in the 3D scene. It improves efficiency but limited by the accuracy of 2D object detection. The LiDAR and RGB features are shared across the backbones in Deep Continuous Fusion \cite{liang2018deep}. MVXNet \cite{sindagi2019mvx} and PointPainting \cite{vora2020pointpainting} fuse the points with the corresponding image features. However, all these approaches use CNNs to fuse the LiDAR and RGB features. In our method, we fuse the RGB and LiDAR features in a novel MSF3DDETR cross-attention block using attention mechanism leveraged by learnt object queries.

\section{MSF3DDETR Network}
\label{sec:MSF3DDETRnetwork}
Our end-to-end 3D object detection network (MSF3DDETR) inputs multi-view RGB and LiDAR data, and outputs 3D bounding boxes for different objects in a scene.

\begin{figure*}[ht]
    \centering
    \includegraphics[width=\linewidth]{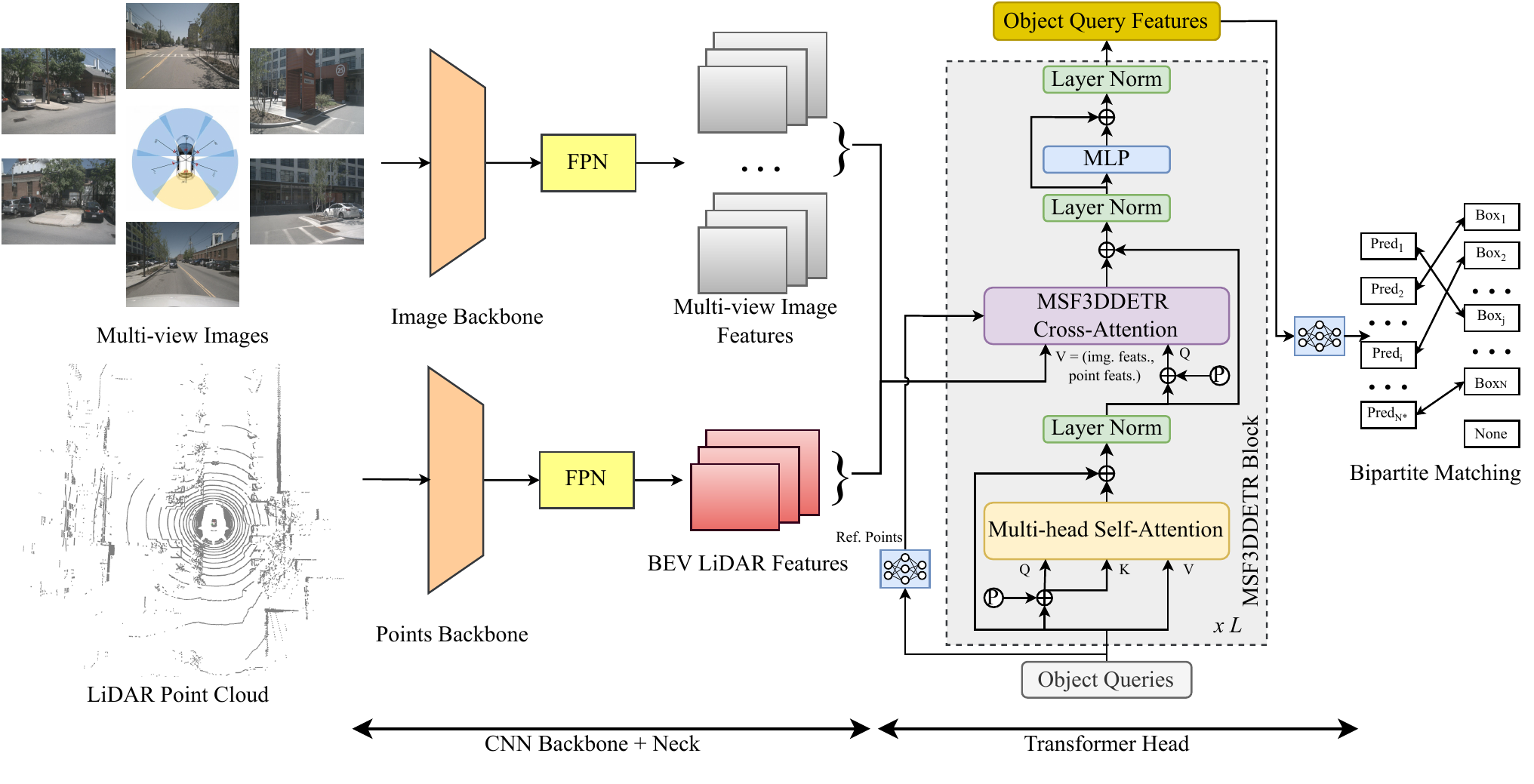}
    \caption{An overview of MSF3DDETR architecture. It is an end-to-end network which inputs multi-view RGB images and LiDAR point cloud to predict 3D bounding boxes. CNN backbone extracts image and BEV LiDAR features which are linked to object queries in the transformer head. Specifically, the novel MSF3DDETR cross-attention block fuses image and LiDAR features by attention mechanism leveraging the learnt object queries.}
    \label{fig:msf3ddetr}
\end{figure*}

Our architecture comprises of three main components as shown in Figure~\ref{fig:msf3ddetr}. Initially, we extract RGB features from multi-view RGB images using a shared ResNet \cite{resnet}, and BEV LiDAR features using VoxelNet \cite{voxelnet} or PointPillars \cite{pointpillars} and SECOND \cite{second} backbone networks. The RGB and BEV LiDAR features are enhanced using a Feature Pyramid Network (FPN) \cite{fpn}. Secondly, a novel transformer head is designed to not only link the computed RGB and LiDAR features with 3D object predictions inspired by \cite{detr3d} but also to efficiently fuse the RGB and LiDAR features by cross-attention mechanism \cite{vaswani2017attention}  leveraging the learnt object queries in the MSF3DDETR cross-attention block. The transformer head starts with object queries which are transformed into 3D object locations, which are projected to the RGB and BEV planes to sample the RGB and BEV LiDAR features via bilinear interpolation. The sampled RGB and BEV LiDAR features are fused with attention weights learnt from object queries. Multi-head self-attention block incorporates fused feature interactions and therby refine object queries. We repeat the MSF3DDETR block multiple times with alternating self-attention, cross-attention and MLP blocks respectively. Finally, we transform the object query features into 3D object parameters and train the network end-to-end with bipartite matching and set-to-set loss \cite{detr}.

\subsection{CNN Backbone and Neck}
Our network inputs multi-view RGB images $\mathcal{I} = \{I_1,\dots, I_6\} \subset \mathbb{R}^{H\times W\times3}$ and LiDAR point cloud $ \mathcal{P} \subset \mathbb{R}^{P\times 3}$. We extract multi-scale, multi-view RGB features using a shared ResNet101 \cite{resnet} backbone with deformable convolutions \cite{dai2017deformable}. The features are further enhanced by FPN to obtain multi-scale features $ \mathcal{F}_1^{img}, \mathcal{F}_2^{img}, \mathcal{F}_3^{img}, \mathcal{F}_4^{img}$, where $ \mathcal{F}_x^{img} = \{f_x^{img1},\dots, f_x^{img6}\} \subset \mathbb{R}^{h\times w\times c} $. To encode the LiDAR point clouds, we use VoxelNet \cite{voxelnet} with 0.1m voxel size or to accelerate object detection for point clouds, we scatter point clouds into BEV pillars using PointPillars \cite{pointpillars} which maps sparse point clouds into dense BEV pillar map. We use SECOND \cite{second} to extract sparse voxel or BEV pillar features and further enhance them using FPN to obtain four sets of features $ \mathcal{F}_1^{bev}, \mathcal{F}_2^{bev}, \mathcal{F}_3^{bev}, \mathcal{F}_4^{bev}$. These features are passed to MSF3DDETR cross-attention block to link with object queries in the transformer head.

\subsection{Transformer Head}
Earlier approaches such as \cite{avod, pillarod, wang2021fcos3d} employs dense set of anchor boxes or produces dense per pixel predictions which results in redundant bounding boxes removed by post-processing methods like NMS. However, NMS-based redundancy removal is non-parallelizable and amounts to increase of inference time. We tackle this issue using a transformer head which uses $ L$ layers of MSF3DDETR block. The MSF3DDETR Block alternates between multi-head self-attention block, MSF3DDETR cross-attention block and MLP block with layer norms and skip connections. 

\begin{figure*}[h]
    \centering
    \includegraphics{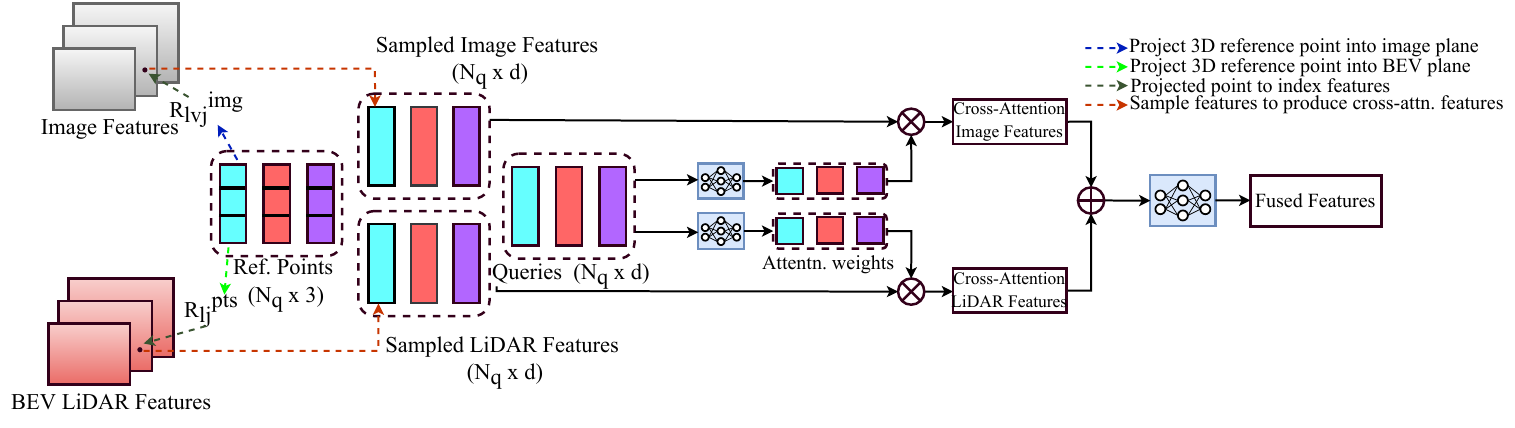}
    \caption{An overview of MSF3DDETR cross-attention block}
    \label{fig:msf3ddetrcross}
\end{figure*}

\noindent\textbf{MSF3DDETR cross-attention} The details are illustrated in Figure~\ref{fig:msf3ddetrcross}. Inspired by DETR \cite{detr}, the transformer head starts with a set of object queries $ \mathcal{Q} = \{q_1,\dots,q_{N_q} \} \subset \mathbb{R}^d$. The queries are transformed to a set of reference points $ \mathcal{R} \subset \mathbb{R}^3$ as:
\begin{equation}
    \mathcal{R} = \phi(\mathcal{Q})
    \label{eq:refpoints}
\end{equation}
where $ \phi$ is a single layer fully connected (FC) network. The reference points are projected to multi-view image planes using sensor transformation matrices to obtain image projected 3D reference points $ \mathcal{R}_{lv}^{img}$ for each layer $ l$ and camera view $ v$. Simultaneously the reference points are also projected to the BEV plane to obtain BEV projected 3D reference points $ \mathcal{R}_l^{pts}$.

The image and BEV features are collected by
\begin{align}
    \begin{split}
        \textbf{f}_{lxv}^{img} &= f^{bilinear}(\mathcal{F}_{xv}^{img}, \mathcal{R}_{lv}^{img}) \\
    \textbf{f}_{lx}^{bev} &= f^{bilinear}(\mathcal{F}_x^{bev}, \mathcal{R}_{l}^{bev})
    \end{split}
\end{align}
where $ \textbf{f}_{lxv}^{img}$ is the image features for $x$-th level of $v$-th camera view at $l$-th layer and $ \textbf{f}_{lx}^{bev}$ is the BEV features for $x$-th level at $l$-th layer.

A neural network is applied on object queries to obtain attention weights for cross-attention of sampled image and BEV features. We further obtain cross-attention image and BEV features which are concatenated and further a neural network is applied to obtain fused image and BEV features as shown in Figure~\ref{fig:msf3ddetrcross}.

The fused features interact with each other in the multi-head self-attention block where the queries are refined iteratively. Finally, we predict bounding box $ \hat{B}_l$ and its class $ \hat{C}_l$ from the refined object queries using a two layer FC networks $ \phi_l^{reg}$ and $ \phi_l^{cls}$, where $ \hat{B}_l = \{\hat{b}_{l1},\dots,\hat{b}_{lj},\dots,\hat{b}_{lN_q}\} \subset \mathbb{R}^9$ and $ \hat{C}_l = \{\hat{c}_{l1},\dots,\hat{c}_{lj},\dots,\hat{c}_{lN_q}\} \subset \mathbb{Z}$. We compute the loss from the predictions for every layer during training but we only use outputs from last layer during inference.

\subsection{Loss}
Inspired by \cite{detr}, we use bipartite matching with hungarian algorithm and set-to-set loss to calculate the error between predicted set $ (\hat{B}, \hat{C})$ and ground-truth set $ (B, C) = (\{b_{1},\dots,b_{j},\dots,b_{M}\}, \{c_{1},\dots,c_{j},\dots,c_{M}\})$. The number of ground-truth boxes $M$ is smaller than number of predicted boxes $N_q$, so we pad the ground-truth boxes with $\emptyset$s (no object) up to $M$. The one-to-one matching between prediction and ground-truth is established with the Hungarian algorithm \cite{kuhn1955hungarian}, $ \sigma^* = \argminG_{\sigma \in \mathcal{P}} \sum_{i=1}^{M}-\mathbb{1}_{\{c_j \neq \emptyset \}}\hat{p}_{\sigma(i)}(c_i) + \mathbb{1}_{\{c_i = \emptyset \}} \mathcal{L}_{box}(b_i, \hat{b}_{\sigma(i)})$, where $\mathcal{L}_{box}$ is the $L_1$ loss for bounding box regression and focal loss \cite{lin2017focal} for class labels. Finally, the set-to-set loss is defined as $ \mathcal{L} = \sum_{i=1}^{N} -\log \hat{p}_{\sigma^*(i)}(c_i) + \mathbb{1}_{\{c_i = \emptyset \}} \mathcal{L}_{box}(b_i, \hat{b}_{\sigma^*(i)})$.

%------------------------------------------------------------------------------------------
\section{Experiments}
\subsection{Dataset}
We evaluate our network on large-scale publicly available nuScenes dataset \cite{caesar2020nuscenes}. The dataset consists of 1K sequences of $\sim$20s duration which are split into 700, 150, 150 sequences for training, validation and testing respectively. The sequences are annotated at 2Hz, leading to 28K, 6K, 6K annotated samples for training, validation and testing respectively. Each sample consists of RGB images from 6 cameras \texttt{[front, front\textunderscore right, front\textunderscore left, back, back\textunderscore left, back\textunderscore right]} and 32-beam LiDAR point cloud with 30K points per sample. 10 categories are available to compute the metrics.

\subsection{Metrics}
We evaluate our model following the official evaluation protocol defined by nuScenes \cite{caesar2020nuscenes}. We evaluate the True Positive (TP) metrics, such as average translation error (ATE), average scale error (ASE), average orientation error (AOE), average velocity error (AVE), average attribute error (AAE), computed in the physical unit. In addition to TP metrics, we also evaluate the main metrics mean average precision (mAP) and the consolidated nuScenes detection score (NDS).

\subsection{Network Architecture}
Our model inputs RGB images from 6 camera views and extract RGB features using a shared ResNet101 \cite{resnet} backbone and FPN \cite{fpn} and thus it produces feature maps of strides 4, 8, 16, 32. We scale the RGB images to 0.8 of input dimensions (1600 $\times$ 900) which achieves a good trade-off between accuracy and efficiency. The LiDAR point cloud features are extracted using SparseConv \cite{sparseconv} to obtain 3D sparse convolutions that refine voxel features. The features are further extracted using a backbone having [5, 5] convolutional layers with dimensions [128, 256]  respectively. The BEV feature maps are combined using a FPN and further a convolutional layer is added at the end to produce four BEV feature maps. The transformer head consists of 6 (layers) MSF3DDETR blocks, where each block has multi-head self-attention, MSF3DDETR cross-attention and MLP with hidden dimension 256. The number of object queries is set to 900.

\subsection{Training Details}
We train our network using AdamW optimizer with an initial learning rate of $2 \times 10^{-4}$. The weight decay is $10^{-2}$. We use Cosine Annealing as learning rate scheduler with a linear warmup until 2K iterations and with minimum learning rate of $2 \times 10^{-7}$. The RGB backbone is initialized with a pretrained FCOS3D \cite{wang2021fcos3d} network and BEV backbone is initialized with a pretrained VoxelNet \cite{voxelnet} network on the same dataset. We train for 24 epochs on two RTX 3090 GPUs with a batch size of 1. We take the top 300 objects with highest category score as final predictions and we do not use any post-processing method such as NMS during inference.

\subsection{Results}
\subsubsection{Quantitative Results}

\begin{table*}[htp]
\centering
\caption{Quantitative results comparison of recent works on nuScenes \cite{caesar2020nuscenes} validation dataset. Modality - C: Camera, L: LiDAR, R: Radar}
\label{tab:quantres}
\begin{tabular}{@{}cccccccccc@{}}
\toprule
Method                & Modality        & NDS $\uparrow$ & mAP $\uparrow$ & mATE $\downarrow$ & mASE $\downarrow$ & mAOE $\downarrow$ & mAVE $\downarrow$ & mAAE $\downarrow$ & NMS \\ \midrule 
% CenterNet \cite{objaspoints}             & C          & 0.328  & 0.306 & 0.716  & 0.264  & 0.609  & 1.426  & 0.658  & \checkmark   \\ \midrule
FCOS3D \cite{wang2021fcos3d}               & C          & 0.415  & 0.343 & 0.725  & 0.263  & 0.422  & 1.292  & 0.153  &  \checkmark   \\ 
DETR3D \cite{detr3d}               & C          & 0.434  & 0.349 & 0.716  & 0.268  & 0.379  & 0.842  & 0.200  &  \ding{55}  \\ \midrule 
% PillarOD \cite{pillarod}              & L           & 0.568  & 0.444 & -      & -      & -      & -      & -      & \checkmark    \\ \midrule
% CenterPoint (pillar) \cite{centerpoint}  & L           & 0.596  & 0.475 & 0.313  & 0.258  & 0.338  & 0.323  & 0.202  & \checkmark    \\ \midrule
% Object DGCNN (pillar) \cite{objectdgcnn} & L           & 0.630  & 0.533 & 0.346  & 0.266  & 0.316  & 0.260  & 0.187  & -    \\ \midrule
CenterPoint (voxel) \cite{centerpoint}  & L           & 0.648  & 0.564 & -  & -  & -  & -  & -  & \checkmark    \\ 
Object DGCNN (voxel) \cite{objectdgcnn} & L           & 0.661  & 0.587 & 0.333  & 0.263  & 0.288  & 0.251  & 0.190  & \ding{55}    \\ \midrule 
CenterFusion \cite{nabati2021centerfusion}         & C+R & 0.449  & 0.326 & 0.631  & 0.261  & 0.516  & 0.614  & \textbf{0.115}  & \checkmark    \\ 
PointPainitng \cite{vora2020pointpainting}        & C+L & 0.581  & 0.464 & -      & -      & -      & -      & -      & \checkmark    \\ 
MVP \cite{mvp} & C+L & \textbf{0.705} &\textbf{0.664} & - & - & - &- & - & \checkmark \\  
MSF3DDETR (ours)                 & C+L & 0.667  & 0.606 & 0.334  & 0.258  & 0.288  & 0.283  & 0.193  & \ding{55} \\ 
MSF3DDETR (ours w/ distillation) & C+L &0.672 &0.613 &\textbf{0.333} &\textbf{0.256} &\textbf{0.287} &\textbf{0.281} &0.191 &\ding{55} \\ \bottomrule
\end{tabular}
\end{table*}

We compare the existing state-of-the-art methods on the nuScenes \cite{caesar2020nuscenes} dataset as shown in Table~\ref{tab:quantres}. Our network surpassed the camera-only CNN and transformer based approaches FCOS3D \cite{wang2021fcos3d} and DETR3D \cite{detr3d} respectively, by approximately 74\% (0.26 mAP), proving our hypothesis that fusion of data improves the accuracy. Our method achieve 4.4\% and 8.6\% improvement in mAP over LiDAR based approaches Object DGCNN \cite{objectdgcnn} and CenterPoint \cite{centerpoint} respectively. The data fusion approaches like CenterFusion \cite{nabati2021centerfusion} combines camera and radar data to produce dense per-pixel predictions, PointPainitng \cite{vora2020pointpainting} annotates each LiDAR point with the CNN features from the camera images and it employs NMS to remove redundant boxes in each view and in overlap regions, and MVP \cite{mvp} employs 2D segmentation and augments virtual points based on 2D segmentation. The result of our camera and LiDAR data fusion approach shows significant improvement in terms of mAP and NDS compared to CenterFusion and PointPainitng without bells and whistles. This observation verifies the significance of our cross-attention based fusion approach. Our method obtains 0.051 less mAP than MVP, but the MVP is a two-stage approach where the prediction result depends on the accuracy of 2D segmentation and it employs NMS to remove redundant boxes which results in significant inference overhead. Moreover, MVP predicts 3D objects for individual multi-view images, whereas our approach predicts the objects in a single-stage by processing multi-view images all at once without NMS. Our network achieve close-to-real-time inference speed (6.3 FPS) on a desktop GPU (RTX 3090), which is approximately 18\% improvement in inference speed over MVP. 

We employ set-to-set knowledge distillation (KD) with a teacher and student model similar to \cite{objectdgcnn} to improve the accuracy. As our approach is NMS-free, we can easily distill the information from teacher to student with homogeneous detection heads. We achieve an increase of 0.007 mAP by KD as shown in Table~\ref{tab:quantres}.

\subsubsection{Qualitative Results}

\begin{figure*}[!h]
    \centering
    \captionsetup[subfigure]{labelformat=empty}
    \begin{minipage}{.3\linewidth}
        \begin{subfigure}{\linewidth}
            \includegraphics[width=\textwidth, height=55pt]{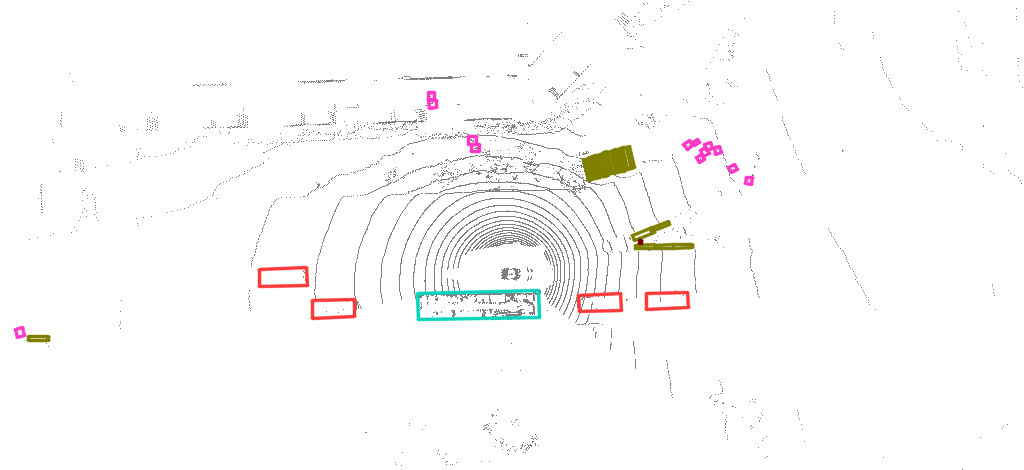}
            \caption{\detokenize{LIDAR_BEV_GT}}
        \end{subfigure}
    \end{minipage}
    \begin{minipage}{0.69\linewidth}
        % \begin{subfigure}{0.30\linewidth}
        %     \includegraphics[width=\textwidth]{images/results/290/v_3_gt.png}
        %     \caption{\detokenize{CAM_FRONT_LEFT_GT}}
        % \end{subfigure}
        % \begin{subfigure}{0.30\linewidth}
        %     \includegraphics[width=\textwidth]{images/results/290/v_1_gt.png}
        %         \caption{\detokenize{CAM_FRONT_GT}}
        % \end{subfigure}
        % \begin{subfigure}{0.30\linewidth}
        %     \includegraphics[width=\textwidth]{images/results/290/v_2_gt.png}
        %     \caption{\detokenize{CAM_FRONT_RIGHT_GT}}
        % \end{subfigure} \\
        \begin{subfigure}{0.30\linewidth}
            \includegraphics[width=\textwidth]{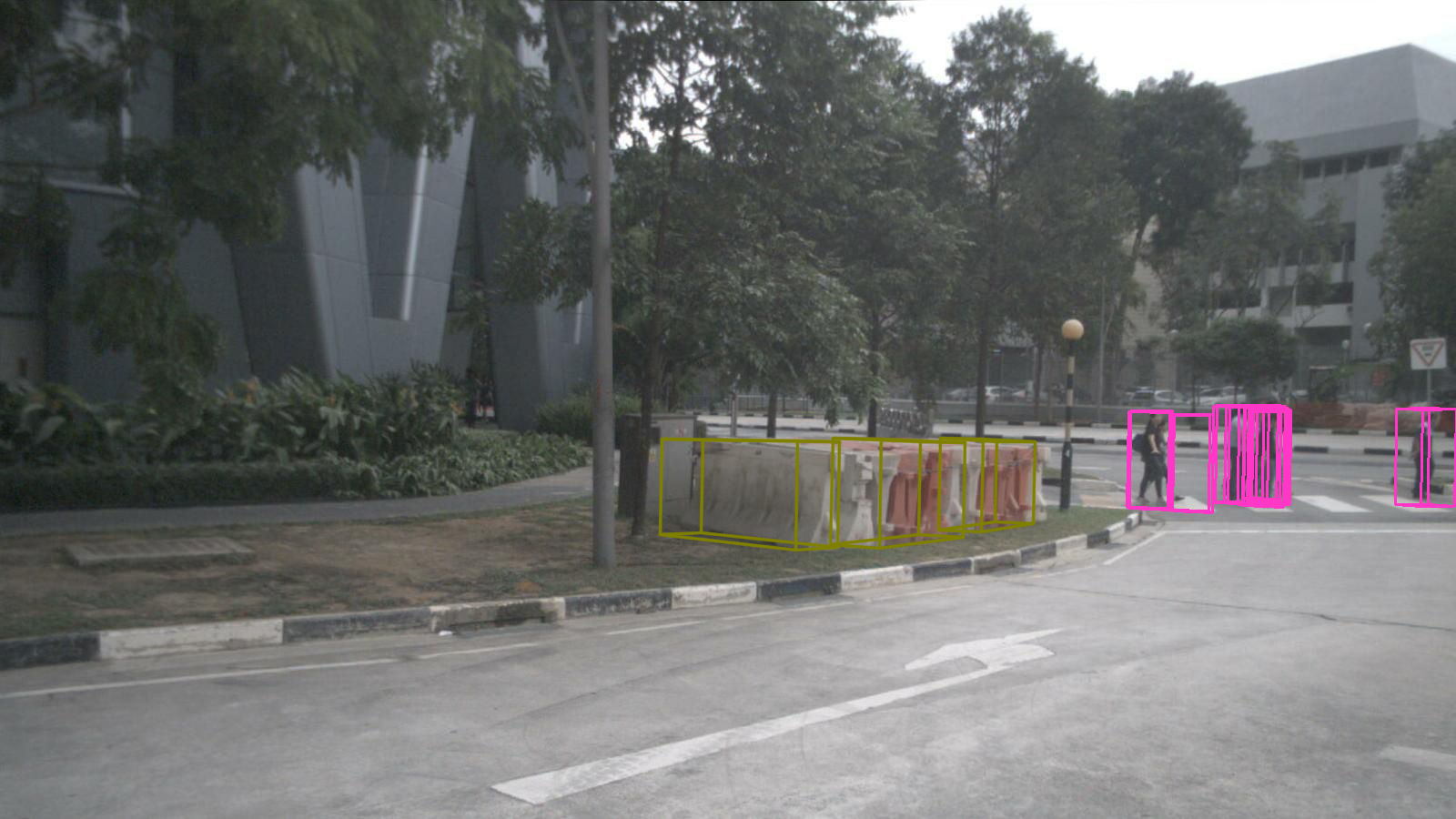}
            \caption{\detokenize{CAM_FRONT_LEFT}}
        \end{subfigure}
        \begin{subfigure}{0.30\linewidth}
            \includegraphics[width=\textwidth]{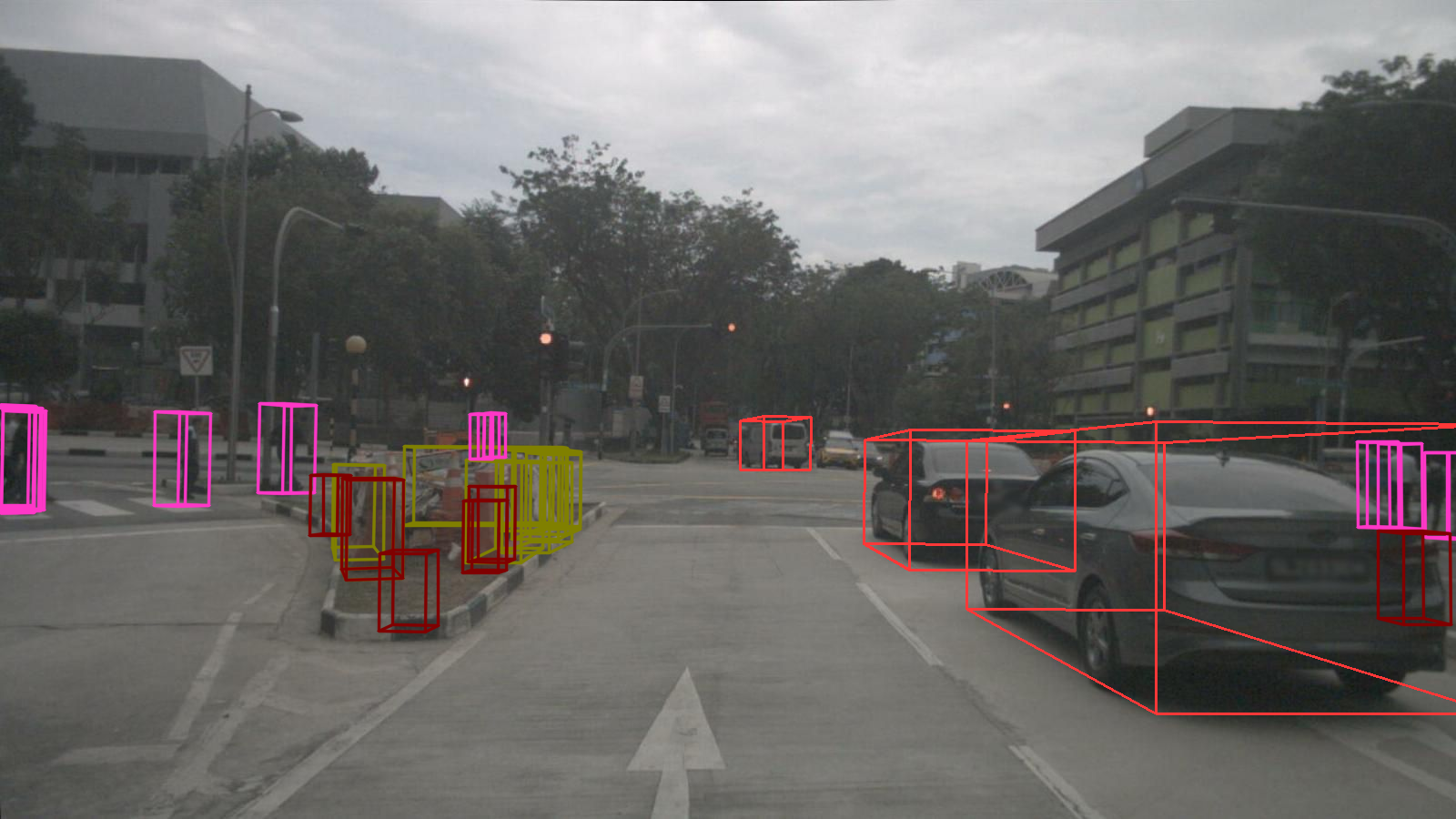}
            \caption{\detokenize{CAM_FRONT}}
        \end{subfigure}
        \begin{subfigure}{0.30\linewidth}
            \includegraphics[width=\textwidth]{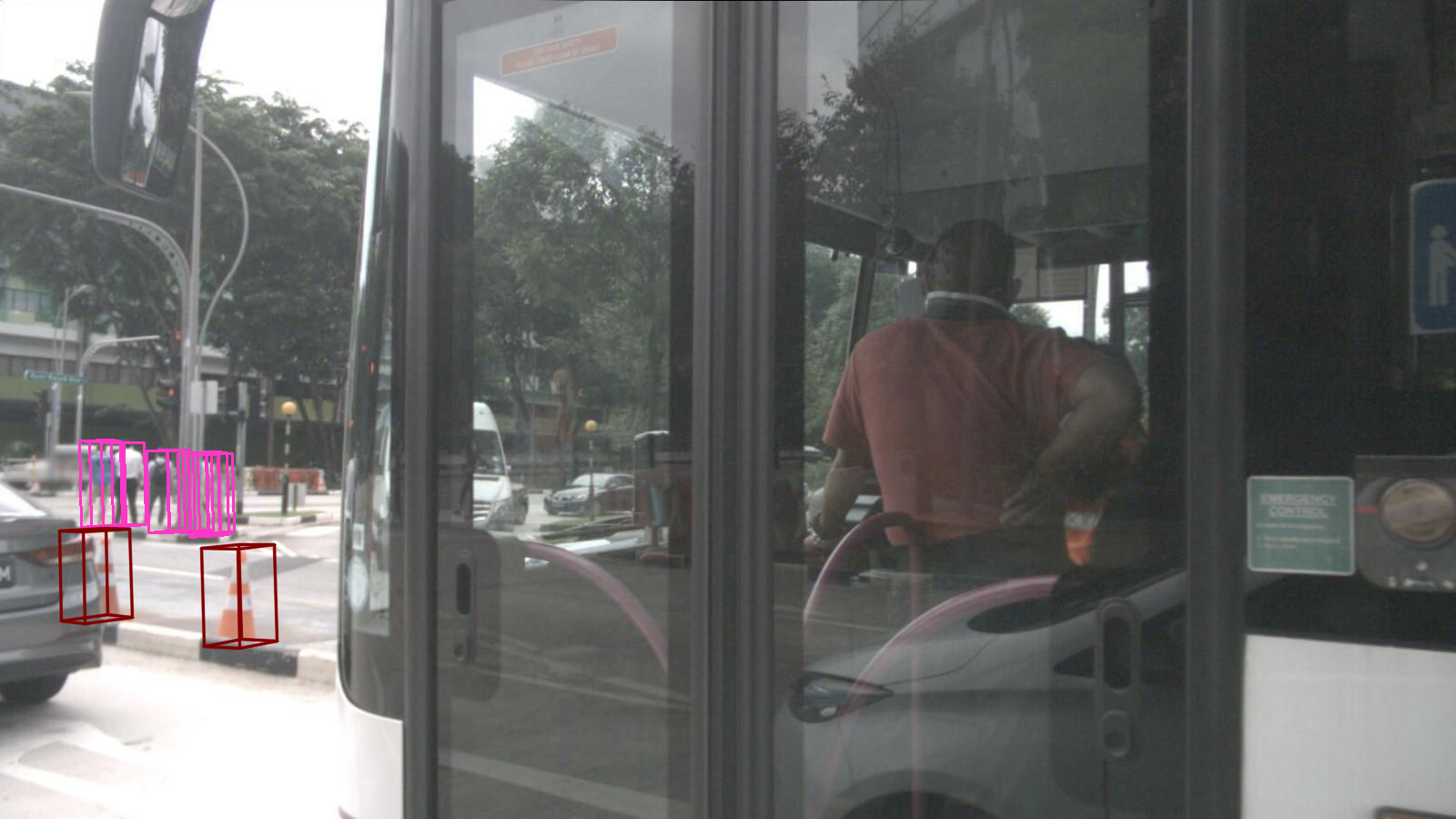}
            \caption{\detokenize{CAM_FRONT_RIGHT}}
        \end{subfigure}
    \end{minipage} \\
    \begin{minipage}{.3\linewidth}
        \begin{subfigure}{\linewidth}
            \includegraphics[width=\textwidth, height=55pt]{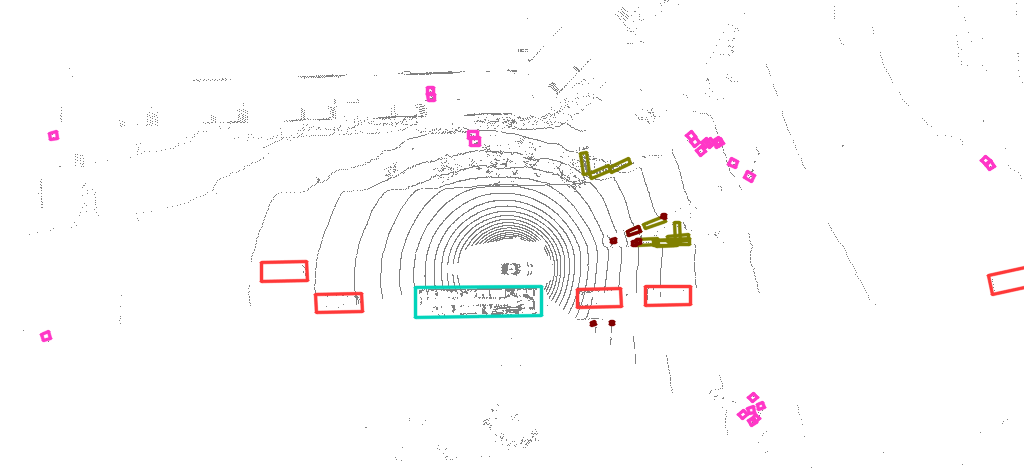}
            \caption{\detokenize{LIDAR_BEV_PRED}}
        \end{subfigure}
    \end{minipage}
    \begin{minipage}{0.69\linewidth}
        % \begin{subfigure}{0.30\linewidth}
        %     \includegraphics[width=\textwidth]{images/results/290/v_5_gt.png}
        %     \caption{\detokenize{CAM_BACK_LEFT_GT}}
        % \end{subfigure}
        % \begin{subfigure}{0.30\linewidth}
        %     \includegraphics[width=\textwidth]{images/results/290/v_4_gt.png}
        %     \caption{\detokenize{CAM_BACK_GT}}
        % \end{subfigure}
        % \begin{subfigure}{0.30\linewidth}
        %     \includegraphics[width=\textwidth]{images/results/290/v_6_gt.png}
        %     \caption{\detokenize{CAM_BACK_RIGHT_GT}}
        % \end{subfigure} \\
        \begin{subfigure}{0.30\linewidth}
            \includegraphics[width=\textwidth]{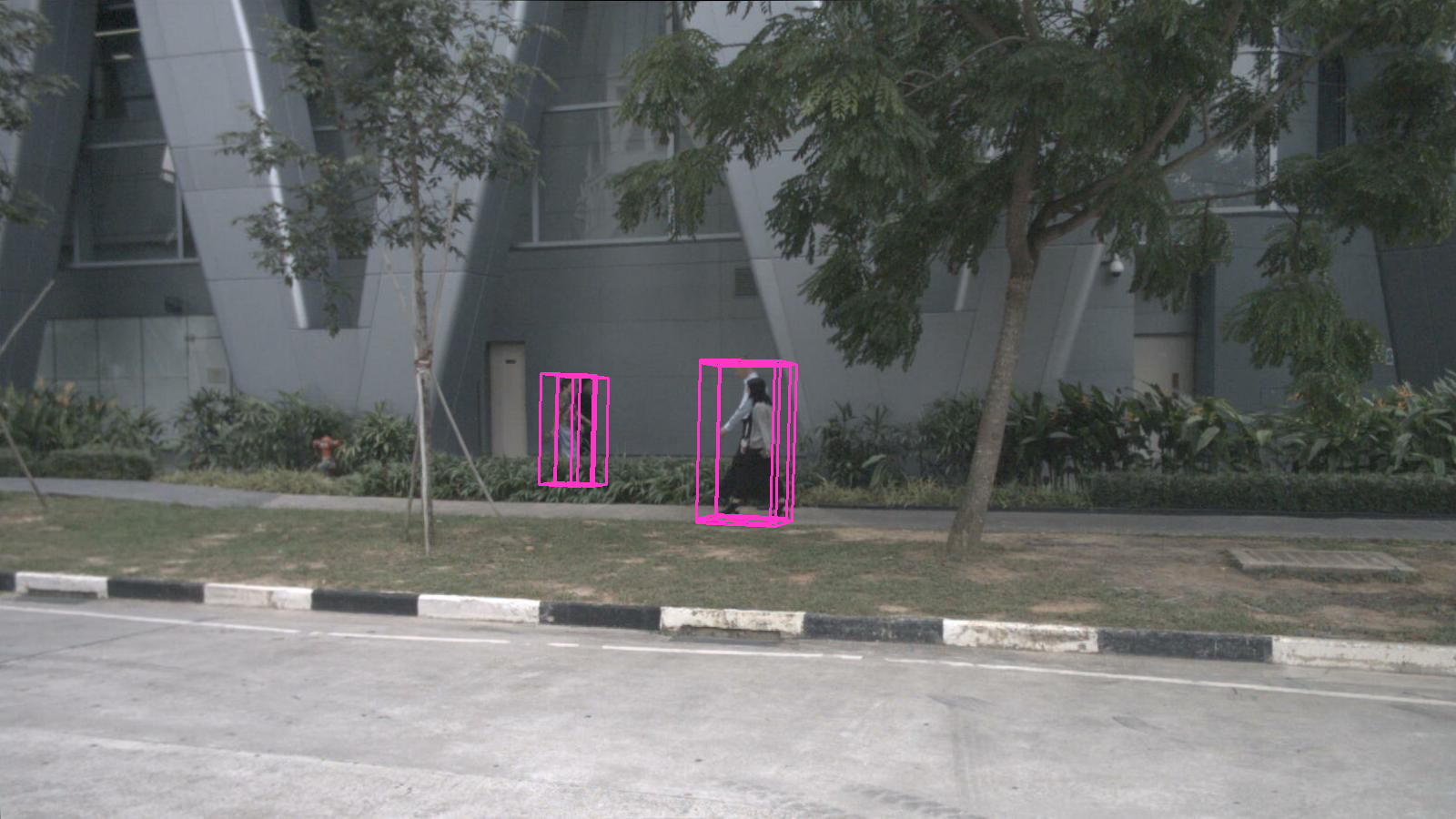}
            \caption{\detokenize{CAM_BACK_LEFT}}
        \end{subfigure}
        \begin{subfigure}{0.30\linewidth}
            \includegraphics[width=\textwidth]{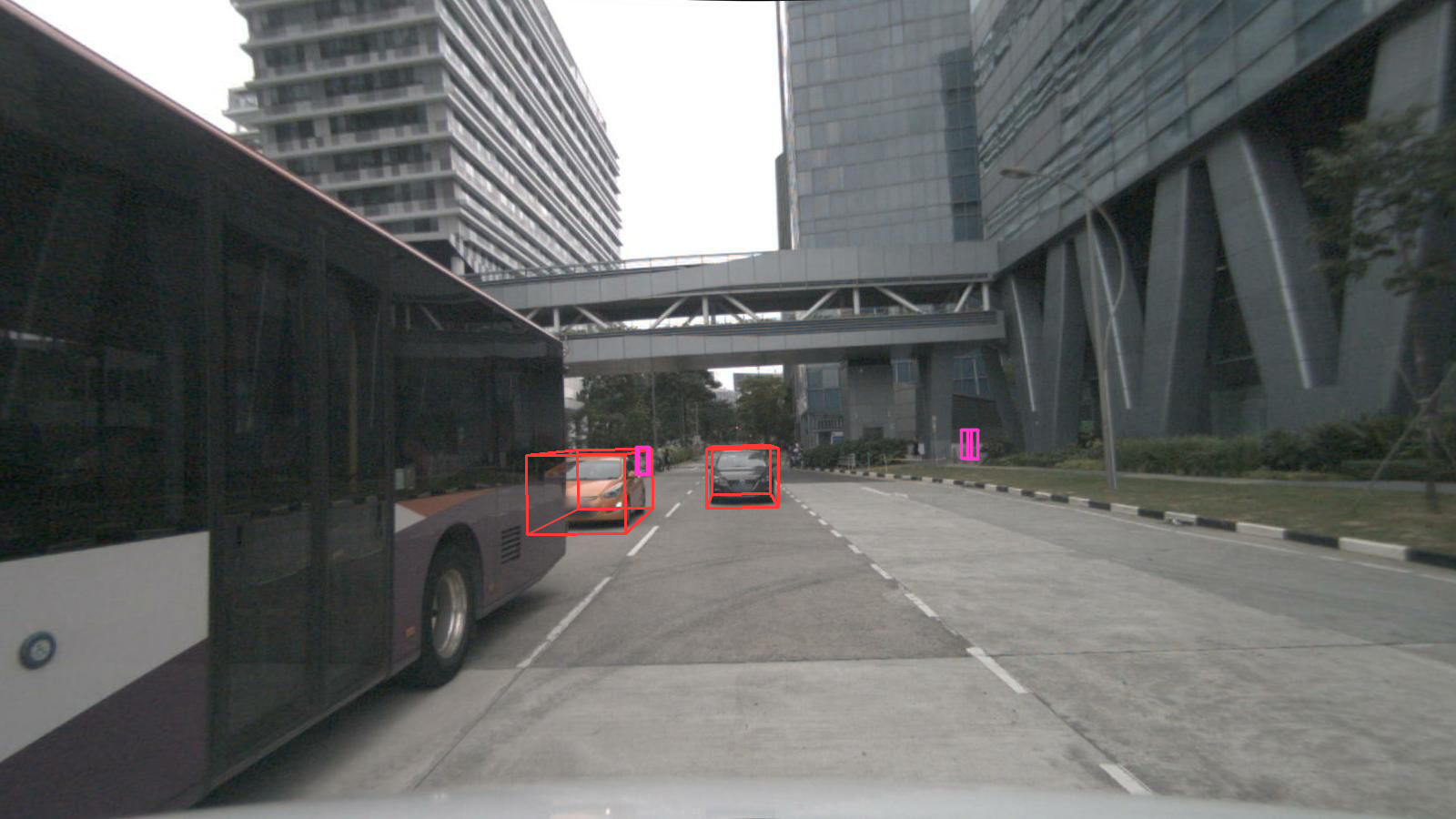}
            \caption{\detokenize{CAM_BACK}}
        \end{subfigure}
        \begin{subfigure}{0.30\linewidth}
            \includegraphics[width=\textwidth]{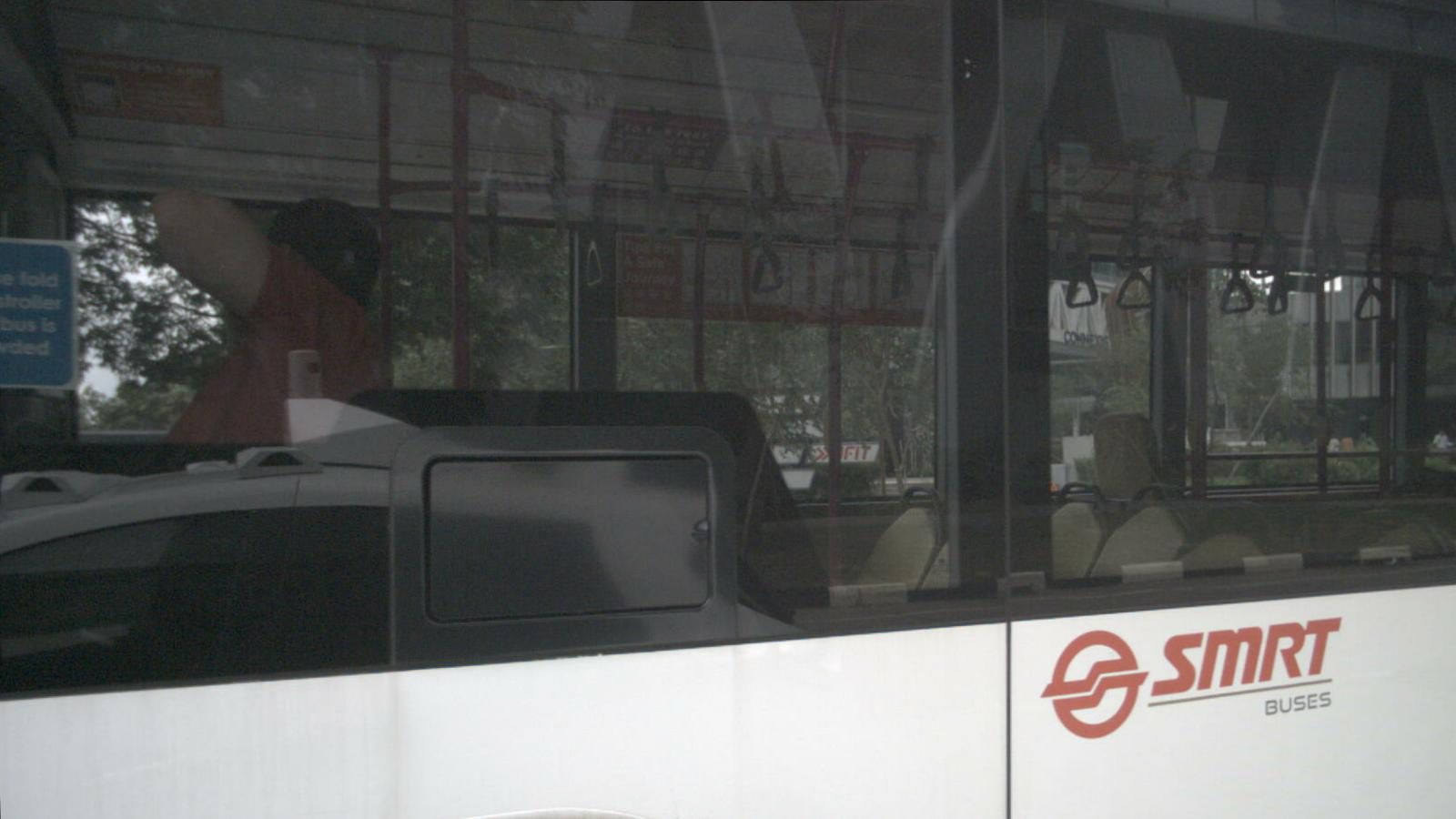}
            \caption{\detokenize{CAM_BACK_RIGHT}}
        \end{subfigure}
    \end{minipage}
    \caption{Qualitative results of MSF3DDETR network both in BEV plane and image planes. Different objects are shown in different colors. Image best viewed in color and zoom.}
    \label{fig:qualtres}
\end{figure*}

The qualitative results of our MSF3DDETR architecture are illustrated in Figure~\ref{fig:qualtres}. The 3D bounding boxes are projected on to the BEV and 6 camera view planes. The MSF3DDETR network was able to detect even the small objects and the objects which are not annotated in the ground-truth, for instance in front of the ego vehicle (right-side of \verb|LIDAR_BEV_PRED|) and also front towards the right (bottom-right of \verb|LIDAR_BEV_PRED|) a car and few persons were not annotated in ground-truth but our network was able to detect them as shown in \verb|CAM_FRONT| and \verb|CAM_FRONT_RIGHT| respectively. Some failure cases include as some barriers in \verb|CAM_FRONT_LEFT| were not detected. A short demo video of the 3D object predictions of MSF3DDETR architecture can be watched at \url{https://youtu.be/F0rxAP1-a24}.

\subsubsection{Analysis}
We analyse the performance of our approach by object categories, object distances and object sizes in comparison to single modality approaches. We report the performance of our approach for different number of layers employed in the transformer detection head.

\textbf{Object category.} We present the Average Precision (AP) of camera-only, LiDAR-only and our fusion based approach for every object category in Table~\ref{tab:apbycat}. LiDAR-only model \cite{objectdgcnn} significantly outperform camera-only model \cite{detr3d} in all the object categories. The camera data helps the LiDAR-only model as in our approach which significantly improves performance on bicycle, bus, motorcycle categories.

\begin{table*}[htp]
\centering
\caption{Average Precision (AP) by object category. CV - Construction Vehicle, Motor - Motorcycle, Ped - Pedestrian, TC - Traffic Cone}
\label{tab:apbycat}
\begin{tabular}{@{}l|llllllllll@{}}
\toprule
Modality        & Car & Truck & Trailer & Bus & CV & Bicycle & Motor & Ped &TC &Barrier \\ \midrule 
Camera-only \cite{detr3d} &0.54 &0.28 &0.16 &0.34 &0.08 &0.27 &0.34 &0.42 &0.52 &0.47 \\
LiDAR-only \cite{objectdgcnn} &0.84 &0.54 &0.40 &0.66 &0.20 &0.44 &0.66 &0.81 &0.64 &0.62 \\
Camera + LiDAR (Ours) &0.86 &0.58 \textcolor{green}{$\uparrow$ 0.04} &0.40 &0.71 \textcolor{green}{$\uparrow$ 0.05} &0.21 &0.53 \textcolor{green}{$\uparrow$ 0.09}  &0.67 &0.83 &0.67 &0.61 \\ \bottomrule
\end{tabular}
\end{table*}

\textbf{Object distance.} The ground-truth bounding boxes are split into three subsets basing on distance of object centers to ego vehicle: \verb|[0m,20m]|, \verb|[20m,30m]| and \verb|[30m,+|$\infty$\verb|]|. Table~\ref{tab:mapbydistance} shows the mAP of different modalities by object distance. The objects that are near to ego vehicle (up to 20 meters distance) there is not much improvement by fusing camera data to LiDAR model, but the performance on objects that are far from ego vehicle (above 30 meters distance) significantly improves by equipping camera data to LiDAR model. This is due to the fact that LiDAR data gets very sparse as the distance from ego vehicle increases.

\begin{table}[htp]
\centering
\caption{mAP by object distance}
\label{tab:mapbydistance}
\begin{tabular}{@{}l|lll@{}}
\toprule
Modality        & \verb|[0m,20m]| & \verb|[20m,30m]| & \verb|[30m,+|$\infty$\verb|]| \\ \midrule 
Camera-only \cite{detr3d} &0.479 &0.257 &0.103  \\
LiDAR-only \cite{objectdgcnn} &0.732 &0.555 &0.303 \\
Camera + LiDAR (Ours) &0.738 &0.575 \textcolor{green}{$\uparrow$ 0.02} &0.333 \textcolor{green}{$\uparrow$ 0.03}\\ \bottomrule
\end{tabular}
\end{table}

\textbf{Object Size.} The ground-truth bounding boxes are split into two subsets based on longer edge of the 3D bounding box: \verb|[0m, 4m]| and \verb|[4m,+|$\infty$\verb|]|. Table~\ref{tab:mapbysize} shows the mAP of different modalities by object size. The performance of our approach on smaller objects (less than 4 meters) is significantly better than larger objects (more than 4 meters). This is due to the fact that camera data has high resolution than LiDAR data which helps in detecting the smaller objects.

\begin{table}[htp]
\centering
\caption{mAP by object size}
\label{tab:mapbysize}
\begin{tabular}{@{}l|ll@{}}
\toprule
Modality        & \verb|[0m,4m]| & \verb|[4m,+|$\infty$\verb|]| \\ \midrule 
Camera-only \cite{detr3d} &0.227 &0.137   \\
LiDAR-only \cite{objectdgcnn} &0.360 &0.254   \\
Camera + LiDAR (Ours) &0.376 \textcolor{green}{$\uparrow$ 0.016} &0.263 \textcolor{green}{$\uparrow$ 0.009}  \\ \bottomrule
\end{tabular}
\end{table}

\textbf{Detection Layers} The quantitative performance of our approach for different layers in the detection head is shown in Table~\ref{tab:detlayers}. The performance of our approach significantly improves as the number of layers in detection head increases, because the object queries are iteratively refined in every layer of the detection head. However, after 6 layers of MSF3DDETR block in detection head there is not much improvement in the performance of our network, so we fix to 6 layers of MSF3DDETR block in the detection head as trade-off between performance and efficiency.

\begin{table*}[htp]
\centering
\caption{Performance by different layers in detection head}
\label{tab:detlayers}
\begin{tabular}{@{}c|ccccccc@{}}
\toprule
Layer &NDS $\uparrow$ & mAP $\uparrow$ & mATE $\downarrow$ & mASE $\downarrow$ & mAOE $\downarrow$ & mAVE $\downarrow$ & mAAE $\downarrow$\\ \midrule
0   &0.607	&0.517	&0.409	&0.273	&0.336	&0.305	&0.187\\ 
1	&0.640	&0.561	&0.359	&0.264	&0.302	&0.289	&0.188\\
2	&0.658	&0.593	&0.340	&0.263	&0.293	&0.287	&0.193\\
3	&0.665	&0.603	&0.336	&0.262	&0.286	&0.285	&0.195\\
4	&0.667	&0.605	&0.335	&0.260	&0.284	&0.285	&0.188\\
5	&0.667	&0.606	&0.334	&0.258	&0.288	&0.283	&0.193\\ \bottomrule
\end{tabular}
\end{table*}

% \begin{figure*}[htp]
%     \centering
%     \captionsetup[subfigure]{labelformat=empty}
%     \begin{minipage}{\linewidth}
%     \begin{subfigure}{0.33\linewidth}
%         \includegraphics[width=\textwidth, height=75pt]{images/results/detlayers/bev_pred_lay1.png}
%         \caption{\detokenize{Layer 1}}
%     \end{subfigure}
%     \begin{subfigure}{0.33\linewidth}
%         \includegraphics[width=\textwidth, height=75pt]{images/results/detlayers/bev_pred_lay2.png}
%         \caption{\detokenize{Layer 2}}
%     \end{subfigure}
%     \begin{subfigure}{0.33\linewidth}
%         \includegraphics[width=\textwidth, height=75pt]{images/results/detlayers/bev_pred_lay3.png}
%         \caption{\detokenize{Layer 3}}
%     \end{subfigure}\\
%     \begin{subfigure}{0.33\linewidth}
%         \includegraphics[width=\textwidth, height=75pt]{images/results/detlayers/bev_pred_lay4.png}
%         \caption{\detokenize{Layer 4}}
%     \end{subfigure}
%     \begin{subfigure}{0.33\linewidth}
%         \includegraphics[width=\textwidth, height=75pt]{images/results/detlayers/bev_pred_lay5.png}
%         \caption{\detokenize{Layer 5}}
%     \end{subfigure}
%     \begin{subfigure}{0.33\linewidth}
%         \includegraphics[width=\textwidth, height=75pt]{images/results/detlayers/bev_gt.png}
%         \caption{\detokenize{Ground-truth}}
%     \end{subfigure}
%     \end{minipage}
%     \caption{Caption}
%     \label{fig:detlayers}
% \end{figure*}

\subsection{Ablation studies}
We report the performance of MSF3DDETR using different number of queries. We show the results for pillar \cite{pointpillars} and voxel \cite{voxelnet} backbones for LiDAR data and ResNet101 \cite{resnet} backbone for camera data. As shown in Table~\ref{tab:queryablation}, the number of queries in the transformer detection head has slight impact on the performance of our approach. However, we use 900 queries in other experiments.

\begin{table}[htp]
\centering
\caption{Ablation study on the number of object queries}
\label{tab:queryablation}
\begin{tabular}{@{}c|cc|cc@{}}
\toprule
 & \multicolumn{2}{|c|}{MSF3DDETR (pillar)} & \multicolumn{2}{c}{MSF3DDETR (voxel)} \\ \midrule
No. of queries & mAP & NDS &mAP &NDS \\ \midrule
300 &0.524 &0.585 &0.587 &0.641 \\
600 &0.536 &0.601 &0.595 &0.652 \\
900 &\textbf{0.545} &\textbf{0.614} &\textbf{0.606} &\textbf{0.667} \\
1000 &0.538 &0.605 &0.591 &0.648 \\\bottomrule
\end{tabular}
\end{table}

We report the performance of MSF3DDETR using different camera-LiDAR backbone combinations as shown in Table~\ref{tab:backboneablation}. We test the performance of MSF3DDETR with ResNet50 and ResNet101 for camera backbones, and PointPillars \cite{pointpillars} with 0.2m pillar size and VoxelNet \cite{voxelnet} with 0.1m voxel size for LiDAR backbones. We can observe from the Table~\ref{tab:backboneablation} that the deep convolutional network ResNet101 with more powerful LiDAR backbone VoxelNet achieves the best performance. However, our approach is more flexible to plugin various backbones for camera and LiDAR data depending on the trade-off between accuracy and efficiency required for the specific application.

\begin{table}[htp]
\centering
\caption{Ablation study on the backbones}
\label{tab:backboneablation}
\begin{tabular}{@{}cc|cc@{}}
\toprule
Camera & LiDAR &NDS &mAP \\ \midrule
ResNet50 & PointPillar 0.2m &0.531 &0.596 \\
ResNet50 & VoxelNet 0.1m &0.595 &0.660 \\
ResNet101 & PointPillar 0.2m &0.545 &0.614 \\
ResNet101 & VoxelNet 0.1m &\textbf{0.606} &\textbf{0.667} \\ \bottomrule
\end{tabular}
\end{table}

\section{Conclusion}
We present MSF3DDETR a single-stage, anchor-free and NMS-free network which inputs multi-view RGB images and LiDAR point clouds to predict 3D bounding boxes. The novel MSF3DDETR cross-attention block links the object queries learnt from the data to RGB and LiDAR features, and thereby helps to fuse the data at object query level, which to the best of our knowledge is the first of its kind. We train on nuScenes dataset and present quantitative and qualitative competitive results to other state-of-the-art approaches. We also employ set-to-set knowledge distillation by teacher and student model to improve the accuracy.

\section*{Acknowledgment}
This work has been supported by European Union's H2020 MSCA-ITN-ACHIEVE with grant agreement No. 765866, Fundação para a Ciência e a Tecnologia (FCT) under the project UIDB/00048/2020 and FCT Portugal PhD research grant with reference 2021.06219.BD.

% \IEEEtriggeratref{1}


\bibliographystyle{IEEEtran}
\bibliography{IEEEabrv,mybibfile}


% \end{thebibliography}

\end{document}